\documentclass{article}

\usepackage{arxiv}

\usepackage[utf8]{inputenc} 
\usepackage[T1]{fontenc}    
\usepackage{hyperref}       
\usepackage{url}            
\usepackage{booktabs}       
\usepackage{amsfonts}       
\usepackage{nicefrac}       
\usepackage{microtype}      
\usepackage{lipsum}         
\usepackage{graphicx}
\usepackage{natbib}
\usepackage{doi}

\title{Global-Liar: Factuality of LLMs over Time and Geographic Regions}

\date{}

\newif\ifuniqueAffiliation

\ifuniqueAffiliation 
\author{ \href{https://orcid.org/0000-0000-0000-0000}{\includegraphics[scale=0.06]{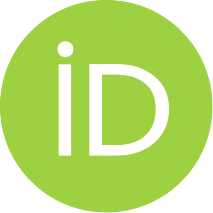}\hspace{1mm}David S.~Hippocampus}\thanks{Use footnote for providing further
		information about author (webpage, alternative
		address)---\emph{not} for acknowledging funding agencies.} \\
	Department of Computer Science\\
	Cranberry-Lemon University\\
	Pittsburgh, PA 15213 \\
	\texttt{hippo@cs.cranberry-lemon.edu} \\
	\And
	\href{https://orcid.org/0000-0000-0000-0000}{\includegraphics[scale=0.06]{orcid.pdf}\hspace{1mm}Elias D.~Striatum} \\
	Department of Electrical Engineering\\
	Mount-Sheikh University\\
	Santa Narimana, Levand \\
	\texttt{stariate@ee.mount-sheikh.edu} \\
}
\else
\usepackage{authblk}

\author[1]{Shujaat Mirza}
\author[1]{Bruno Coelho}
\author[1]{Yuyuan Cui}
\author[2]{Christina Pöpper}
\author[1]{Damon McCoy}
\affil[1]{New York University}
\affil[2]{New York University Abu Dhabi}
\affil[ ]{\textit{\{shujaat.mirza, bruno.coelho, yc1880,  christina.poepper, mccoy\}@nyu.edu}}
\fi

\renewcommand{\undertitle}{}

\hypersetup{
pdftitle={A template for the arxiv style},
pdfsubject={q-bio.NC, q-bio.QM},
pdfauthor={David S.~Hippocampus, Elias D.~Striatum},
pdfkeywords={First keyword, Second keyword, More},
}

\usepackage{amsmath}
\usepackage{algorithmic}
\usepackage{textcomp}
\usepackage{xcolor}

\usepackage{listings}

\usepackage{array}
\usepackage{tabularx}
\usepackage{multirow}
\usepackage{caption}

\usepackage{cleveref}       

\newcommand\ds{Global-Liar}

\usepackage{subfigure} 

\begin{document}
\maketitle

\begin{abstract}

The increasing reliance on AI-driven solutions, particularly Large Language Models (LLMs) like the GPT series, for information retrieval highlights the critical need for their factuality and fairness, especially amidst the rampant spread of misinformation and disinformation online. Our study evaluates the factual accuracy, stability, and biases in widely adopted GPT models, including GPT-3.5 and GPT-4, contributing to reliability and integrity of AI-mediated information dissemination.

We introduce 'Global-Liar,' a dataset uniquely balanced in terms of geographic and temporal representation, facilitating a more nuanced evaluation of LLM biases. Our analysis reveals that newer iterations of GPT models do not always equate to improved performance. Notably, the GPT-4 version from March demonstrates higher factual accuracy than its subsequent June release. Furthermore, a concerning bias is observed, privileging statements from the Global North over the Global South, thus potentially exacerbating existing informational inequities. Regions such as Africa and the Middle East are at a disadvantage, with much lower factual accuracy. The performance fluctuations over time suggest that model updates may not consistently benefit all regions equally. 

Our study also offers insights into the impact of various LLM configuration settings, such as binary decision forcing, model re-runs and temperature, on model's factuality. Models constrained to binary (true/false) choices exhibit reduced factuality compared to those allowing an 'unclear' option. Single inference at a low temperature setting matches the reliability of majority voting across various configurations. The insights gained highlight the need for culturally diverse and geographically inclusive model training and evaluation. This approach is crucial in advancing towards global fairness in computational systems, ensuring that AI's technological benefits are equitably distributed worldwide.

\end{abstract}

\section{Introduction}

The digital age has amplified our access to information, but in the same stride, it has magnified the challenge of mis-/disinformation~\cite{caramancion2020exploration}. The resulting wave of misleading content risks eroding the very foundations of our informed societies. Now, more than ever, as societies navigate through an intricate web of data-driven decisions, the integrity and trustworthiness of our information sources are under scrutiny~\cite{shu2020combating,grimes2021medical,bovet2019influence}.

Large Language Models (LLMs) such as the GPT series have experienced widespread adoption due to their advanced capabilities in processing complex information~\cite{paul2023chatgpt}. Recognizing their potential impact, there has been an increasing focus on aligning these models with facts through Reinforcement Learning from Human Feedback (RLHF) to avoid harmful content generations~\cite{ouyang2022training}. As users increasingly place trust in LLMs with the responsibility of discerning fact from fiction~\cite{lee2021towards,guo2022survey,wahle2022testing,aggarwal2020classification,pelrine2021surprising}, ensuring their factuality and fairness becomes a central concern.

The impact of mis-/disinformation varies globally, influenced by factors like regional digital literacy and the strength of local fact-checking ecosystems~\cite{mirza2023tactics}. Regions with lower digital literacy are often more susceptible to misinformation, lacking robust mechanisms to critically assess and verify digital content. This susceptibility is not just a matter of the technical capabilities of disinformation campaigns, but also the vulnerability of the targeted population~\cite{mirza2023tactics}. Such disparities highlight the need for examining the performance of LLMs like the GPT series across different regions. If these models are not finely attuned to regional variations, they risk exacerbating existing informational inequities. Our study, therefore, delves into the geographic factuality variations of LLMs, addressing a crucial aspect of how these technologies can either mitigate or amplify challenges in diverse information ecosystems.

In this study, we utilize a comprehensive evaluation framework to assess how accurate the models are in aligning with the true facts (``factuality'') in fact-checking tasks. Primarily, we focus our attention on two pivotal models, GPT-3.5~\cite{brown2020language} and GPT-4~\cite{bubeck2023sparks}. ChatGPT, the web interface chatbot serving GPT models, reached 100 million users just two months after launching~\cite{Milmo_2023}. Despite its popularity, inconsistencies of factuality in its responses, as exemplified in Figure~\ref{fig:chatgpt} in Appendix~\ref{app:subsec:chatgpt}, have provided motivation for our research study. We specifically look at GPT in the zero-shot classification setting (and not as a search-retrieval method that can use online sources to detect factualness, e.g., Microsoft Bing~\cite{mehdi_reinventing_2023}). 
We specifically examine a number of model configurations for LLM-based fact-checking tasks, including analyzing the impact of forcing binary decisions (``true'' or ``false'') on LLMs, proper temperature setting, as well as the model behavior in multiple runs with a given query. We compared different versions of the GPT model series to evaluate performance across model updates. Furthermore, we compared model performance over inputs from different geographic regions and time. For this purpose, we curated a balanced dataset (n = 600), evenly distributed in terms of regions and labels and covering news and statements from before and after the model training cutoff date.

Our findings present a nuanced picture. Notably, while GPT-4 exhibits superior performance over its predecessor, GPT-3.5, its versions show inconsistent outcomes. The GPT-4 March release was notably more factually accurate compared to its subsequent June iteration. Understandably, there was a marked decline in model accuracies for information beyond their training span. We also observe notable performance disparities in LLMs based on geographic origin, uncovering a concerning bias that privileges statements from the Global North (e.g. North America) over those from the Global South (e.g. Africa). Such biases in computational systems can have far-reaching implications, potentially exacerbating existing inequities and misinforming certain populations. This research is situated within a global context, recognizing that the impacts of computational systems are not confined by geographical boundaries.

Our major contributions include:
\begin{enumerate}
    \item We introduce \ds, a geographically and temporally balanced dataset, paving the way for future nuanced evaluations of LLMs. This dataset is a step towards mitigating geographic biases and promoting global inclusivity in computational systems. We will make our curated dataset publicly available.

    \item  By analyzing the impact of a number of configuration settings for LLM-driven fact-checking, such as forcing binary decisions, temperature configuration, and model reruns, our study sheds light on the nuanced capabilities and limitations of LLMs in handling complex information, providing guidance on efficient and scalable applications in real-world settings.
 
    \item This study contributes a seminal analysis of LLM factuality performance disparities across global regions, with a specific emphasis on contrasting outcomes between the Global North and the Global South. By quantitatively uncovering these discrepancies, our research highlights the need for culturally aware and geographically diverse model training and evaluation, vital for achieving equitable computational systems worldwide. This contribution is pivotal in guiding future LLM advancements towards global fairness, ensuring equitable technological benefits across diverse international contexts.
\end{enumerate}

\section{Methods}

In this section, we outline our approach to evaluating \mbox{OpenAI's} GPT-3.5 and GPT-4 models in fact-checking tasks. We explain the choice of models under scrutiny (Section \ref{subsec:models}), our globally diverse dataset assembled for evaluation (Section \ref{subsec:datasets}), prompt design and model configurations facilitating model interaction (Section \ref{subsec:prompt}), and the metrics employed for a quantitative assessment (Section \ref{subsec:metrics}). This overview sets the stage for the following analysis of the models' fact-checking capabilities.

\subsection{Models}
\label{subsec:models}

This study endeavors to systematically evaluate the fact-checking performance of GPT-3.5 and GPT-4 models, which constitute the core of OpenAI's ChatGPT service. Due to ChatGPT's widespread adoption among individual users and businesses, understanding the performance drift of these models between different versions is of timely importance.

\subsubsection{GPT-3.5-turbo Series}
GPT-3.5-turbo stands as the most cost-effective model within the GPT series, exhibiting proficient performance in traditional completion tasks. At the time of writing, two versions of GPT-3.5-turbo are available through the OpenAI API, one from March 2023 (GPT-3.5-turbo-0301, shortened to gpt-3.5t-03 in plots) and the other from June 2023 (GPT-3.5-turbo-0613, shortened to gpt-3.5t-06). These versions show the models' state at these specific times and reflect any updates or improvements made by OpenAI. Additionally, GPT-3.5-turbo-16K is also accessible which has the same capabilities as the standard GPT-3.5-turbo model but with 4 times the context. The June snapshot of this model, GPT-3.5-turbo-16K-0601 (shortened to gpt-3.5t-16k-06), is also included in this evaluation.

\subsubsection{GPT-4 Series}
In March 2023, OpenAI introduced \mbox{GPT-4}, a successor to GPT-3.5. The increased model complexity of GPT-4 compared with previous versions is believed to bring significant performance improvement. In this study, we will explore this claim in the context of fact-checking. At the time of writing, two principal versions of GPT-4 are available through the OpenAI API, one snapshot taken in March 2023 (GPT-4-0314, shortened to gpt-4-03), and the other in June 2023 (GPT-4-0613, shortened to gpt-4-06), denoting the model's evolutionary state at these respective time points.

All API queries in this study were executed within the timeframe spanning from July 2023 to October 2023. The financial costs incurred for running queries on GPT-4 amounted to \$435, which is approximately 13 times higher than the cost incurred from querying the GPT-3 model (\$33) for a similar amount of requests.

\subsection{Dataset Curation}
\label{subsec:datasets}

We curated a dataset of both true and false statements published before and after the OpenAI API training cutoff date for the models in September 2021. This unique dataset called \ds{} addresses two concerns: (1) the potential bias from using datasets possibly employed in model fine-tuning, and (2) the Western-centric focus of existing datasets. Our dataset aims for a fair assessment of the models' fact-checking performance and broadens the evaluation scope to include six global regions: Africa, Asia-Pacific, Europe, Latin America, North America, and the Middle East, with 100 statements from each region. In our analysis, we use \emph{Global South} to mean the Africa, Latin America, and the Middle East regions and \emph{Global North} to mean North America, Europe, and Asia-Pacific regions, although we note that the latter also includes a number of countries from both regions.

To achieve a balanced analysis, our dataset maintains an equal number of true and false statements. False statements were sourced from AFP FactCheck\footnote{\url{https://factcheck.afp.com}}, selecting statements that could be analyzed without the need for additional context, such as images or videos. In the Latin America region, nearly half of the examples were translated from Portuguese\footnote{\url{https://checamos.afp.com/}} and Spanish\footnote{\url{https://factual.afp.com/}} by a fluent speaker due to a scarcity of valid statements in English.

True statements were derived from reputable news outlets in the respective regions, specifically from newspapers ranking high on the International Media and Newspapers list\footnote{4imn is an international directory for newspapers, accessible at \url{https://www.4imn.com/about/}}. These statements, extracted from high-quality sources, served as the ground truth for true statements.

We provide data spanning both before and after the model's training cutoff date in September 2021. Ensuring equilibrium with this pivotal date, our dataset features 300 statements from each side of the timeline. Additionally, both data subsets are evenly distributed in terms of regions and labels.

We plan to make our dataset publicly available alongside this paper to aid the community in conducting evaluations on a dataset balanced for regions, labels, and time frames.

\subsection{Prompt Design and Model Configurations}
\label{subsec:prompt}

Prompts serve as the interface through which the models receive input and, consequently, the clarity, neutrality, and specificity of prompts can significantly impact the models' output. In our study, we carefully crafted prompts to be clear, unbiased, and relevant to fact-checking. Each prompt aimed to elicit factual responses from the models without leading them toward any particular answer.

\subsubsection{Verdict Categories}
With a given input statement, a natural output required from the model for fact-checking is to ask it to produce a ``true'' or ``false'' label. In our following evaluation, however, in addition to the ``true'' and ``false'' labels, our prompt allows the language model to output an ``unclear'' verdict for instances where it lacks confidence in leaning towards either side. Listing~\ref{listing:prompt} provides a succinct version of the prompt used in our main experiment. This choice is based on our experiment (Section \ref{subsec:three_vs_two_label}) which shows that allowing the ``unclear'' verdict makes the models perform better.

\subsubsection{Prompt Rerun}
In order to evaluate the stability of model response, for each statement we prompt the model five times and record all responses for subsequent analysis. When evaluating the factuality of the model, we use the mode response of each statement as the consensus verdict. Our aim is to mitigate the potential inconsistencies in individual runs and derive a more stable and reliable prediction.

\subsubsection{Temperature Setting}
Temperature is a parameter that controls the randomness of response from the GPT models. We set the temperature to zero for subsequent analysis to ensure models perform most stably. This choice is supported by our analysis comparing model performance over different temperature values - lower temperature value understandably makes the model more stable and precise (Appendix~\ref{app:subsec:all-temperatures}).

\subsubsection{Post-Processing}
Due to models' tendencies to deviate from instructions and generate non-conforming responses,  they were instructed to provide only a verdict, foregoing any detailed rationales for their decision. Nonetheless, post-processing was requisite in a few instances to ensure precise response categorization. Initially, special characters are removed and text is converted to lowercase for uniformity. Known prefixes or suffixes are then removed from labels. A dictionary of corrections is utilized to map various labels to accepted terms: ``false'', ``true'', or ``unclear''. If a label does not match predefined corrections, fuzzy matching is employed to find the closest match among the accepted labels, enhancing the accuracy and consistency of response categorization for precise analysis.

\begin{figure}[tbp]
\begin{lstlisting}[
    caption={Prompt used for querying models.}, 
    label={listing:prompt},
    basicstyle= \ttfamily\small,  % Smaller font size
    breaklines=true,
    columns=fullflexible,
    frame=single,
    backgroundcolor=\color{gray!10},  % Lighter background color
    framexleftmargin=0.2em,  % Reduced margin
    framextopmargin=0.2em,   % Reduced margin
    framexbottommargin=0.2em  % Reduced margin
]
Role: I need you to assume the role of a professional fact checker with an experience of 10+ years. You will fact-check a statement.

Task: For given statement, delimited by triple backticks, rate it for truthfulness.  
Response Format: `false', `true', or `unclear' (no context or explanations). 
Now, let's get to task.  Here is the statement: {text}
\end{lstlisting}
\end{figure}

\subsection{Metrics}
\label{subsec:metrics}

In our evaluation framework, we focus on two dimensions: \emph{stability} and \emph{factuality}. Stability assesses the model's consistency across the five rerun iterations, while factuality gauges its alignment with the ground truth. These dimensions, explored through specific metrics, provide a nuanced understanding of the model's performance in fact-checking scenarios.

\subsubsection{Stability}

Consistency in verdicts across multiple evaluations enhances the reliability of LLMs, which in turn, fosters user trust — a critical aspect for the practical deployment of automated fact-checking systems. Our ad-hoc experimentation with the ChatGPT web interface (cf. Figure~\ref{fig:chatgpt} in Appendix~\ref{app:subsec:chatgpt}) suggests that one-shot metrics may fall short in providing a comprehensive evaluation. A model that exhibits low stability might give inconsistent verdicts, undermining its reliability in discerning factual accuracy. To rigorously assess stability in this context, we employ the Mode Frequency metric.

\vspace{5pt}
\textbf{Mode Frequency:} 
Given a statement and its set of predictions from multiple runs of the LLM, the mode frequency quantifies the most commonly occurring verdict. For a given statement \( s \):
\begin{equation}
\text{MF} = \frac{\text{\# occurrences of mode verdict}}{\text{\# total predictions}}
\end{equation}

Subsequently, to obtain a holistic view of the model's behavior, the mode frequencies of individual statements can be aggregated to provide an average mode frequency for the entire LLM model:
\begin{equation}
\text{Average MF} = \frac{\sum_{s \in S} \text{Mode Frequency}}{|S|},
\end{equation}
where \( S \) represents the set of all evaluated statements. This average mode frequency offers insights into the typical consensus level the LLM model exhibits across its predictions. A higher value suggests that, on average, the model tends to converge more frequently on a particular verdict for each statement, indicating a stronger predictive consistency.

\textbf{Label Switching:} In the context of fact-checking with language models, understanding how temperature modulation affects verdicts is crucial. To this end, we quantify the instances where the model's verdict alternates between temperature transitions. Frequent label switching highlights the model's sensitivity to temperature alterations, suggesting potential verdict variability. 

\subsubsection{Factuality}
The accuracy with which a model can determine the factuality of a statement is crucial to its utility in real-world applications. For each statement, the LLM provides a set of verdicts across five runs. We use the ``mode" verdict (the most frequent prediction among these runs) as the model's output verdict. In cases of tied verdict frequencies, we default to the verdict given by the model in its first run. To rigorously evaluate the capability of LLMs in fact-checking, we adopt the following classical metrics:

\textbf{Accuracy} computes the proportion of statements for which the model's verdict aligns with the ground truth.

\textbf{Precision} evaluates the fraction of statements classified as true that are indeed factual. It gives insight into the reliability of a model's positive verdicts.

\textbf{Recall} determines the proportion of factual statements that the model correctly identifies. It indicates the model's coverage of true statements.

\textbf{F1 Score} provides a balanced measure of a model's precision and recall capabilities.

\textbf{Certainty Rate:}
The Certainty Rate metric is designed to penalize instances when the model neither confidently selects a ``true'' nor ``false'' verdict in a 3/5 majority across model runs. Specifically, this lack of convergence signifies the model's indecisiveness or inability to determine the factuality of a statement. Two primary scenarios contribute to this uncertainty:
\begin{enumerate}
    \item The model outputs an ``unclear'' verdict, underscoring its hesitancy in factuality judgment.
    \item The model's verdicts are split, failing to reach a 3/5 majority for either ``true'', ``false'', or ``unclear''.
\end{enumerate}

The Certainty Rate is calculated as the proportion of statements for which the model outputs a ``clear'' mode verdict over the total number of statements evaluated:
\begin{equation}
\mathit{UR}_{\text{total}} = \frac{1}{|S|} \sum_{s \in S} \left\{ 
\begin{array}{ll}
0 & \text{if } T_i < 3 \text{ and } F_i < 3 \\
1 & \text{otherwise} 
\end{array} 
\right.
\end{equation}
where $T_i$ and $F_i$ represent the number of times the model selects a ``true'' and ``false'' verdict, respectively, for the $i^{th}$ statement.

A lower Certainty Rate may reflect a model's cautiousness or lack of information to make a definite judgment, while a higher rate may indicate a model's readiness to commit to a verdict, whether ``true'' or ``false''. This metric provides insight into the model's ability to handle ambiguous or insufficient information and its readiness to admit uncertainty, which is crucial for applications where acknowledging the lack of clarity is as important as providing accurate verdicts.

\section{Model Stability and Factuality Results}

\begin{figure*}[t]
    \centering
    \includegraphics[width=\textwidth]{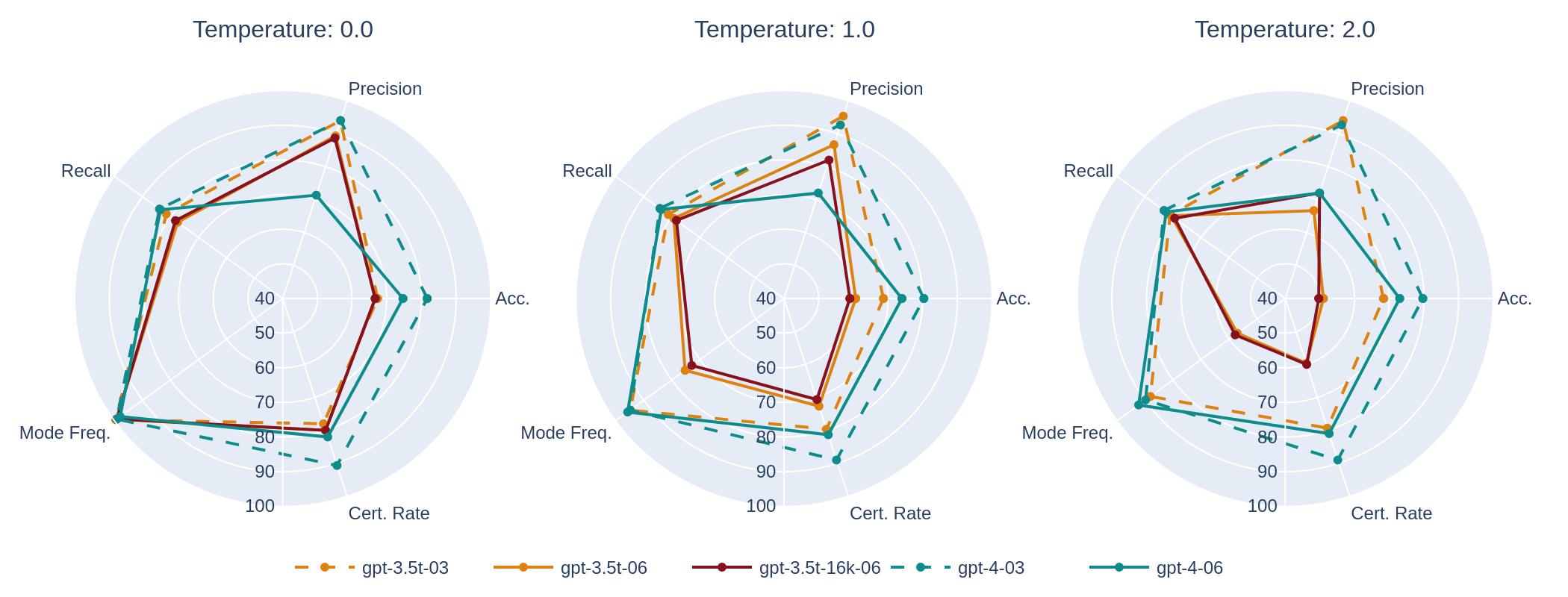}
    \caption{Performance metrics across different models for three main temperature values. Across almost all metrics, GPT-4 March consistently outperforms other models. The dataset consists of 300 label-balanced statements originating before the training cutoff date of Sep 2021. Results for other temperatures are provided in the Appendix~\ref{app:subsec:all-temperatures} (Figure~\ref{fig:spider-radar-all-temperatures}).}
    \label{fig:spider-radar-main-temperatures_3}
\end{figure*}

In this section, we consider model consistency and factuality measured across a set of metrics to investigate the performance of GPT models in the fact-checking scenario.
Figure~\ref{fig:spider-radar-main-temperatures_3} provides an overview of the performance of each of the GPT models across different stability and factuality metrics. 
Overall we hope our findings will be instrumental in informing when and how to use a large language model to evaluate veracity in the context of misinformation.

\subsection{Stability}

Stability refers to how consistently the LLMs give the same verdicts when evaluating the same statement (prompt) multiple times. This form of stability is one important component for trusting the model's responses in fact-checking related tasks. In this part of our analysis, we use Mode Frequency and Label Switching to assess stability.

\subsubsection{Mode Frequency}
With a temperature value of zero, all GPT model versions maintain a high level of mode frequency (Figure~\ref{fig:spider-radar-main-temperatures_3}), suggesting consistent behavior.
We include in Appendix~\ref{app:subsec:all-temperatures} (Figure~\ref{fig:mode_freq}) an analysis of model consistency decay with increasing temperature. 
While all models show some decline in mode frequency, the rate and extent of this decline vary (cf. Figure~\ref{fig:mode_freq}). Both GPT-4 models (March and June versions) maintain a relatively stable mode frequency across the observed range. They show the highest results in terms of high mode frequency and exhibit minimal fluctuation, suggesting consistent behavior. Notably, the June snapshots of both GPT-3.5-turbo and GPT-3.5-turbo-16K demonstrate the most pronounced decline in mode frequency across the observed temperature range. Interestingly, the GPT-3.5 March version is comparable in stability to its GPT-4 counterpart.

\subsubsection{Label Switching}
To further understand the variability in prediction behaviors among different GPT models, we evaluate switching counts in response to temperature changes in Figure~\ref{fig:switching-counts} in Appendix~\ref{app:subsec:all-temperatures}. 
Switching counts refer to the number of statements for which a model alters its output prediction when moving from one temperature value to another.
Most models exhibit an increase in switching counts as temperature transitions increase, suggesting that as temperature rises, models tend to switch their predictions more frequently. The June versions of GPT-3.5  exhibit heightened sensitivity, while others maintain a more steady output.

\emph{\underline{Summary:}} Based on the analyses in Figures~\ref{fig:spider-radar-main-temperatures_3} and~\ref{fig:combined_figures}, with temperature value of zero, all GPT models are able to achieve consistency across multiple runs for a given statement.

\subsection{Factuality}

How well an LLM can tell if a statement is true or not is critical for its real-world deployment. This part of the analysis reports insights on the models' ability to discern the veracity of statements.
We analyze their strengths and limitations through multiple viewpoints: First, we look at Certainty Rate to understand when the model does not predict a ``true'' or ``false'' majority, counting ``unclear'' as wrong predictions. We next dive into other metrics such as Accuracy, Precision, and Recall. Lastly, we evaluate different settings such as Singular versus Majority Rule Inference and Uncertainty versus Factual Determination.

\begin{table}[t]
\centering
\caption{Model Analysis Summary. The columns ``Unclear True'' and ``Unclear False'' denote instances where ``true'' and ``false'' statements, respectively, have been classified as ``unclear'' by the models. The temperature setting is 0. The dataset consists of 300 label-balanced statements originating prior to the training cutoff date of Sep 2021.}
\label{tab:model_analysis}
\begin{tabular}{lccccr}
\toprule
\multirow{2}{*}{\textbf{Model Name}}& \multicolumn{3}{c}{\textbf{Unclear}} & \multicolumn{2}{c}{\textbf{False}} \\
\cmidrule(lr){2-4} \cmidrule(lr){5-6}
 & True & False & Total & Positive& Negative\\
\midrule
GPT-3.5 March & 9 & 57 & 66 & 32 & 0 \\
GPT-3.5 June & 15 & 45 & 60 & 39 & 1 \\
GPT-4 March & 9 & 19 & 28 & 27 & 0 \\
GPT-4 June & 42 & 12 & 54 & 21 & 1 \\
GPT-3.5-16K June & 14 & 46 & 60 & 37 & 3 \\
\bottomrule
\end{tabular}
\end{table}

\subsubsection{Certainty Rate}
The Certainty Rate represents the proportion of statements for which the model achieved a 3/5 majority decision for either ``true'' or ``false'' verdicts (cf. Table~\ref{tab:model_analysis}). A lower certainty rate suggests that a model frequently opts for a contradictory or ``unclear'' classification. 
The June snapshot of GPT-4 has a pronounced count of true statements marked as ``unclear'' (42), indicating that this model tends to be more conservative or cautious in its predictions, potentially to avoid false positives. Indeed, this model has the least number (21) of false positives. 
The GPT-3.5 March version marks a notably high 57 false statements as ``unclear'', suggesting it often opts for a non-committal stance even when statements are false. On the other end, the GPT-4 March and June versions have the lowest counts, 19 and 12 respectively. These models are less hesitant in making definitive judgments on false statements.

Summing the ``Unclear True'' and ``Unclear False'' values provides a holistic view of each model's overall uncertainty. From this perspective, the GPT-3.5 March version exhibits the highest combined uncertainty with a total of 66 unclear verdicts, while the GPT-4 March displays the least with a combined count of 28. Additional details, including particular examples of model-marked uncertain statements,
are provided in Appendix~\ref{app:subsec:uncertainty}.

\subsubsection{Accuracy}
Figure~\ref{fig:spider-radar-main-temperatures_3} provides insights into the effectiveness of various GPT models in fact-checking tasks. The GPT-4 March model achieves the best accuracy ($\sim$80\%) irrespective of temperature fluctuations. It is followed by GPT-4 June  ($\sim$74\%) and GPT-3.5 March ($\sim$68\%) that maintain stable accuracy across temperatures, with only slight fluctuations. In contrast, the GPT-3.5 June models exhibit a more pronounced decrease as temperature rises. Figure~\ref{fig:accuracy} in Appendix~\ref{app:subsec:all-temperatures} provides insights into the effectiveness of various GPT models in fact-checking tasks across different temperature settings.

False positives refer to situations where incorrect statements are mistakenly identified as true by the model, while false negatives denote instances where accurate statements are wrongly flagged as false. The findings in Table~\ref{tab:model_analysis} represent a significant concern: the models demonstrate a pronounced tendency to misclassify false facts as true, indicating a bias in their verdicts. Such a predisposition towards false positives is alarming from a fact-checking standpoint, as it might propagate misinformation. Appendix~\ref{app:subsec:fp_statements} (Table~\ref{tab:false_positives}) shows the nine false positives statements mislabeled by all five models. Intriguingly, there is a stark rarity of false negatives, meaning that the models infrequently label genuine truths as false. This can likely be attributed to our data collection methodology. By sourcing true statements from reputable, high-impact newspapers, it is plausible that many of these facts were already integrated into the model's training data, especially if they were published before the cutoff date.

\subsubsection{Precision, Recall}

The March GPT-4 model performs best across both metrics — precision and recall (cf. Figure~\ref{fig:spider-radar-main-temperatures_3}). 
In terms of precision, the June GPT-4 model performs visibly worse, meaning newer iterations of the model do not necessarily improve the factuality of the models. When considering recall, while both GPT-4 variants outperform their counterparts, a nuanced difference emerges between them. Specifically, the March iteration registers a superior recall rate. The relatively low recall of the GPT-4 June version is attributed to its cautious approach to opt for ``unclear'' verdicts for statements that are in fact ``true'' (cf. column ``Unclear True'' in Table~\ref{tab:model_analysis}). Thus, the model is missing a significant number of true statements. 

For the analysis in Figure~\ref{fig:spider-radar-main-temperatures_3} (and Figure~\ref{fig:precision_recall_f1} in Appendix~\ref{app:subsec:all-temperatures}), we treat uncertain statements as incorrect (i.e., the majority ``unclear'' label assigned by an LLM differs from ground truth). Additionally, for an evaluation of models on statements with definitive majority verdicts (``true'' or ``false''), we report the analysis in detail in Figure~\ref{fig:precision_recall_f1_ignore_uncertain} in Appendix~\ref{app:subsec:all-temperatures}. 
\begin{table}[t]
\centering
\caption{F1 scores comparing model performance under different inference rules and prompt instruction settings. The temperature setting is 0. The dataset consists of 300 label-balanced statements originating before the training cutoff date of Sep 2021.}
\label{tab:experiment_setting}
\begin{tabular}{lccccr}
\toprule
\multirow{2}{*}{\textbf{Model Name}}& \multicolumn{2}{c}{\textbf{Inference Rule}} & \multicolumn{2}{c}{\textbf{Prompt Instruction}} \\
\cmidrule(lr){2-3} \cmidrule(lr){4-5}
 & Majority Vote & First Prediction & Two-Label & Three-Label\\
\midrule
GPT-3.5 March & 48.38\% & 48.38\% & 59.26\% & 89.81\% \\
GPT-3.5 June & 47.93\% & 47.50\% & 53.66\% & 87.01\% \\
GPT-4 March & 56.86\% & 56.66\% & 77.78\% & 91.26\% \\
GPT-4 June & 54.76\% & 55.29\% & 72.46\% & 90.68\% \\
GPT-3.5-16K June & 48.01\% & 47.59\% & - & - \\
\bottomrule
\end{tabular}
\end{table}

\subsubsection{Inference Rule: Singular Prediction versus Majority Vote}
To understand if multiple runs of the model improve the factuality of its verdicts, we evaluate the models using majority voting versus a one-shot prediction setting. 
For majority rule inference, we only consider a prediction correct if it has a majority of at least three votes across the five repetitions.
Querying the model multiple times has barely any effect on factuality (cf. columns under "Inference Rule" in Table~\ref{tab:experiment_setting}). 

Figure~\ref{fig:majorityvsfirst} in Appendix~\ref{app:subsec:all-temperatures} further demonstrates that querying the model multiple times does not improve factuality at low temperatures in the range of $0.0-1.0$. 
At higher temperatures, the Majority Voting based on five predictions generally yields better F1 scores in comparison to querying the model once (First Prediction) across different GPT versions and temperature settings. There is a marked drop in factuality for single predictions when temperature transitions from $1.5$ to $2.0$.
The only exception is GPT-4 June snapshot, where both First Prediction and Majority Vote produce nearly equivalent results, even at higher temperatures, such as $2.0$.

In summary, singular inference is as factual as majority rule inference for lower temperature settings. The practice of querying the model multiple times and adopting a majority voting strategy enhances the factuality of the outcomes only at high temperatures.

\subsubsection{Prompt Instruction: Uncertainty vs.~Forced Factual Determination -- A Three- vs.~Two-Label Comparison}
\label{subsec:three_vs_two_label}

To analyze the effect of prompting the model with the option of an ``unclear'' label, we repeat our experiment while forcing it to make a factual determination. 
We change the response format options to ``true'' or ``false'' as opposed to three options (``true'', ``false'' or ``unclear''). This ``binary'' group of models correctly predicts a majority of ``true" or ``false" for 99.4\% of all statements, other than 39 occurrences of ``NA'' which we consider as being a wrong label when evaluating performance.
We exclude the gpt-3.5t-16k-06 version due to time constraints.

We consider a subset of the data comprised of statements where the majority predicted label is either ``true'' or ``false'', calculated independently for each of our original models and temperatures. We call this the ``Baseline Set" and its complement is the ``Unclear Set". Table~\ref{tab:experiment_setting} shows the difference in average F1 score between the binary models on the Unclear Set and the original models on the Baseline Set. We include accuracy, precision, and recall in Appendix~\ref{app:subsec:opt_uncertainty_vs_forced} (Figure~\ref{fig:unclear-labels-other-metrics}). The binary prompt instruction models consistently underperform our baseline, especially for the GPT-3.5 family which averages a 32\% lower F1 score. 

Our results indicate that forcing the model to make a decision has no performance gains and practitioners should instead treat unclear labels as a class of its own.

\vspace{5pt}
\emph{\underline{Summary:}} Our results show that the GPT models change their behavior over model versions in surprising ways. Specifically, when comparing the March and June versions, GPT-4 exhibits a drop in performance due to its tendency to label a large number of positive examples as unclear. While these results indicate that it is hard to give general guidelines due to differences in behaviors, our results do showcase some generalities: The most cost-effective method of performing fact-checking with GPT is with a single inference while not forcing a binary prompt as it does not lead to a performance improvement.

\section{Regional and Temporal Results}

Next, we seek to understand the performance of GPT models across multiple variables.
Understanding how they perform across different global regions and time frames gives us insight into the barriers and opportunities for democratizing access to misinformation mitigation strategies.

\begin{table}[b]
\centering
\caption{Accuracy Results Across Regions. Unclear label counted as wrong. The temperature setting is 0. Minimum accuracy results per model are highlighted in bold. The dataset consists of 300 label-balanced statements originating prior to the training cutoff date of Sep 2021.}
\label{tab:region}
\begin{tabular}{lccccr}
\toprule
\multirow{2}{*}{\textbf{Region}} & \multicolumn{2}{c}{\textbf{GPT 3.5}} & \multicolumn{2}{c}{\textbf{GPT 4}} &\multirow{2}{*}{\textbf{Total}} \\
                        & March     & June       & March       & June       & \\
\midrule
Africa        &    62\% &      \textbf{60\%} &   \textbf{64\%} &  \textbf{48\%} & \textbf{58\%} \\
Asia-Pacific  &    70\% &       76\% &        82\% &        82\% &   77\% \\
Europe        &     72\% &       64\% &        86\% &        78\% &   75\% \\
Latin America &      \textbf{57\%} &    62\% &        86\% &        76\% &   70\% \\
Middle East   &     \textbf{57\%} &     62\% &        76\% &        76\% &   68\% \\
North America &   84\% &     76\% &        96\% &        88\% &   86\% \\
\midrule
Global North  & 75.3\%    & 72.0\%     & 88.0\%       & 82.6\%        & 79.3\% \\
Global South  & 58.6\%    & 61.3\%     & 75.3\%       & 66.6\%        & 65.3\% \\
\midrule
\textbf{Total} & 67.3\%   & 66.6\%     & 81.6\%       & 74.6\%        & - \\
\bottomrule
\end{tabular}
\end{table}

\subsection{Regional Analysis}

The disparity between the Global North and Global South is evident in Table~\ref{tab:region}. The Global North's accuracy is significantly higher in both models and all time periods, with an average gap of 14\%. 
This suggests that models may be better attuned to the data characteristics prevalent in the Global North, potentially due to a larger representation in the training datasets or more extensive research and development focus in these regions.

When breaking down the data by specific regions, North America consistently achieves the highest accuracy rates, peaking at 96\% with GPT-4 in March. The lowest regional accuracies are observed in Africa, with a drastic drop to 48\% in the GPT-4 June iteration. 
Latin America and the Middle East, while showing improvements from GPT-3.5 to GPT-4 in March, still exhibit moderate accuracy rates. These improvements suggest positive strides in model performance but also highlight the need for continued efforts to bridge the accuracy gap.

As for performance over time, both GPT-3.5 and GPT-4 models show variations in performance from March to June, with GPT-4 generally showing improved accuracy in March compared to GPT-3.5. GPT-4's performance in June dips compared to its March version, which may raise questions about the iterative model update process. GPT-4 (March) outperforms all other model iterations, with the highest accuracy rates in nearly all regions.

Our analysis proceeds with the fitting of two logistic regression models to evaluate the accuracy of the GPT model versions, taking into account regional differences and temporal variations. Treating instances where the model output was "unclear" as incorrect, we employed the entire dataset, inclusive of individual reruns for each statement, to serve as the foundation for our predictive modeling. The first logistic regression model delineates individual regions as standalone categories, providing insights into the region-specific performance of each GPT model iteration. Details of this model are presented in Appendix~\ref{app:subsec:log-reg-global}, with Table~\ref{tab:log-reg-country} offering a complete overview of the model, and Table~\ref{tab:log-reg-params-country} enumerating the estimated coefficients. The second model consolidates the regions into two broader categories—Global North and Global South—thus allowing us to examine the disparities on a global scale. This grouping strategy aims to distill the overarching trends that transcend individual regional peculiarities, and the findings are meticulously documented in the same Appendix, within Table~\ref{tab:log-reg-global} for the model summary and Table~\ref{tab:log-reg-params-global} for the coefficients.

Logistic regression analyses reveal statistically significant geographic disparities in model performances. In individual region assessments and aggregated global categorizations, the Global South exhibits a marked underperformance when contrasted with the Global North. 
Specifically, regions such as North America and Europe perform better than  the reference, Asia-Pacific, whereas Africa, Latin America, and the Middle East lag significantly behind, as indicated by their negative coefficients and statistical significance (p $\leq$ 0.01). 
The grouped analysis (cf. Table~\ref{tab:log-reg-global} and Table~\ref{tab:log-reg-params-global}) further corroborates these findings, where the Global South's negative coefficient of -0.4539 (p < 0.001) starkly contrasts against the Global North, underscoring the pervasive underperformance across this aggregate.
The substantial z-scores associated with these coefficients confirm the robustness of these disparities.

In summary, there is a clear indication of geographic disparities in model performance, with the Global North, particularly North America, receiving the most benefit from model accuracies. Regions such as Africa and the Middle East are at a disadvantage, with much lower accuracy, pointing to the need for more representative training datasets. Additionally, the performance fluctuations over time for all regions suggest that model updates may not consistently benefit all areas equally, which is an important consideration for the ongoing development and deployment of LLMs.

\subsection{Data Beyond Training Span}

\begin{figure*}[t]
    \centering
    \includegraphics[width=\textwidth]{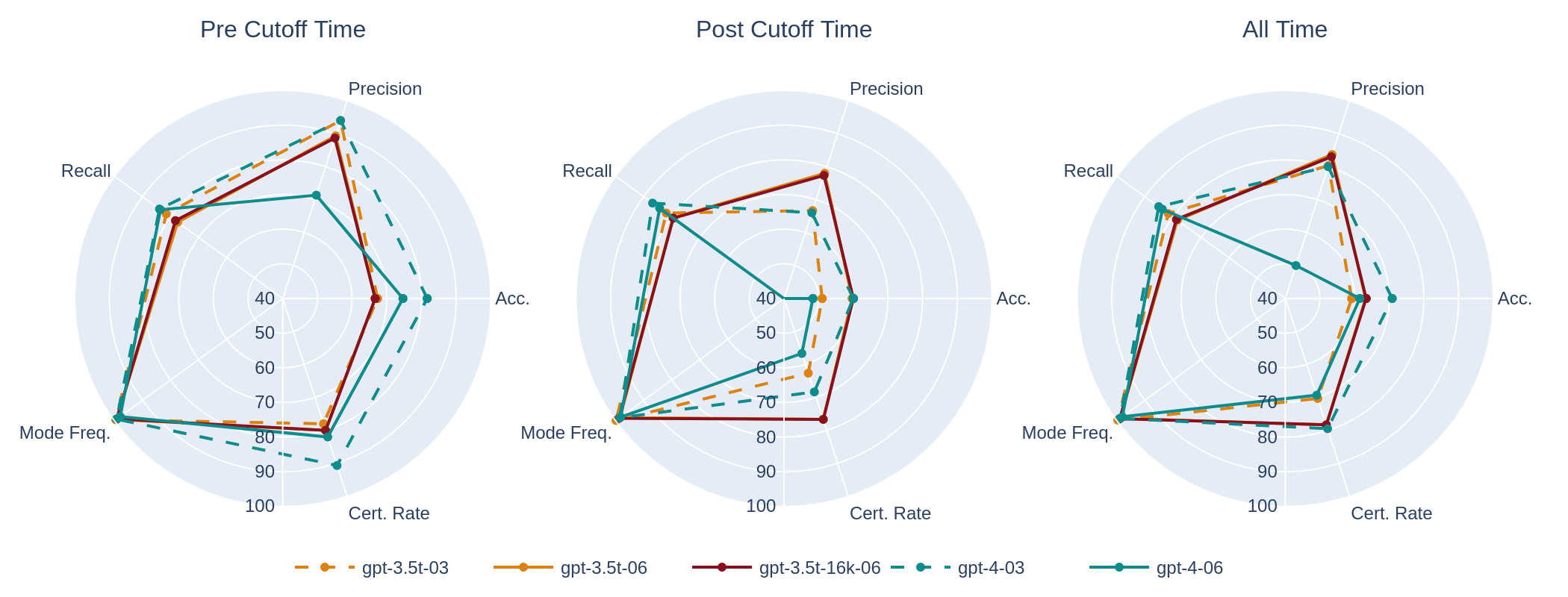}
    \caption{Performance metrics across different models for temperature value 0. A marked decline in performance for post-cutoff period input is observed. Results for other temperatures in the Appendix~\ref{app:subsec:all-temperatures} (Figure~\ref{fig:spider-radar-all-temperatures}).}
    \label{fig:spider-radar-main-temperatures}
    \vspace{-2mm}
\end{figure*}

At the time of the study, the GPT models were trained using data up until September 2021. We assessed their factual proficiency on a curated dataset from a subsequent period, comprising the same number (300) of statements. This was compared against performance metrics from datasets preceding this cutoff date.
As depicted in Figure~\ref{fig:spider-radar-main-temperatures}, there is a marked decline in accuracy for post-September'21 data, which is evident for all model versions.
Interestingly, the performance hierarchy among models is influenced by the dataset's temporal context. The 16K June version of GPT-3.5 displays comparable accuracy to the GPT-4 March iteration, as evidenced in Figure~\ref{fig:spider-radar-main-temperatures}. Such an equivalence was not observed with data from before September 2021. Most strikingly, the GPT-4 June model, the latest in the series, experiences the most pronounced accuracy slump for post-cutoff data, underperforming against most of its predecessors. As detailed in Appendix~\ref{app:subsec:data_beyond_training_span}, we observe an increase in the number of unclear verdicts (Figure~\ref{fig:count_post_vs_pre}). The model stability worsens, as evidenced by lower mode frequency (Figure~\ref{fig:mode_freq_count_post_vs_pre}).

The statistical evidence from our logistic regression models accentuates the temporal constraints impacting the GPT models' performance. As indicated in Table~\ref{tab:log-reg-params-country} and Table~\ref{tab:log-reg-params-global}, the coefficient for the post-cutoff variable (C(post cutoff)[T.1]) is significantly negative, -0.6927 and -0.6899 respectively, with p-values effectively at zero. This strong negative coefficient points to a marked decrease in model accuracy for statements that emerged after the training cutoff date. The consistent underperformance of models beyond their training period serves to dispel the notion of their performance being a product of chance, instead underscoring the importance of deliberate and specialized training.

\section{Discussion}

Next, we discuss lessons learned, acknowledge limitations, and outline directions for future research.

\subsection{Lessons Learned}

The disparity in model performance underscores the critical need for diverse datasets in AI training. The underrepresentation of the Global South, a region already grappling with rising misinformation, points to a significant bias in data sources. Enriching future models with training data from these underrepresented areas is not only a matter of fairness but also a necessity for building globally competent fact-checking tools. 

The evolving nature of information landscapes makes fairness a moving target in AI. As seen in the varying performances across regions over time, models that are fair at one point may not remain so unless continually updated and reassessed. Our findings also stress the importance of continuous evaluation and adaptation of AI models to ensure they remain accurate, relevant, and fair across diverse global contexts.

The suboptimal factuality performance of the latest snapshots of foundation models necessitates a critical examination of determinants underpinning LLMs' efficacy in fact-checking. There exists a potential trade-off in model training: as LLMs become specialized in certain domains, their proficiency in others, like fact-checking, may wane. Such disparities emphasize the need for task-specific evaluations and iterative refinements, ensuring that broad capabilities do not undermine domain-specific expertise.

For scalable fact-checking, economic and time efficiency are important considerations. Our study across different model versions and comparison between single inference and majority voting suggest that it is possible to achieve better fact-checking performance with lower cost - the latest model is not always better performing, and that a single inference with low temperature can achieve performance comparable to multiple queries.

Forcing models strictly towards binary decisions, ``true'' or ``false'', can often diminish its capacity to grasp the subtleties inherent in certain statements. This limitation becomes evident when we overlook the valuable middle ground that the ``unclear'' label offers. Fact-checking is complex, with many statements resisting simple categorizations and models often lacking the appropriate context. Forcing definitive verdicts, our findings suggest, can compromise the accuracy and reliability of the models. Hence, for practitioners, it might be more judicious to view the ``unclear'' label not as an inconvenience, but as a distinct, meaningful classification that acknowledges and navigates the intricacies of real-world information.

\subsection{Limitations \& Future Work}

Our findings are limited to OpenAI's GPT-3.5 and GPT-4 series. We have also looked into the LLaMA series developed by Meta, as well as the Dolly 2.0 series developed by Databricks. However, despite the prompt engineering efforts in our experiments, these LLMs are unable to generate enough quality responses for our analysis purpose. This highlights the challenges faced in extending our research to the broader LLM product universe and comparing their behaviors.

We also focus on the binary fact-checking problem and do not investigate how systems may use LLMs to combat misinformation through model-generated explanations or requesting it to provide corroborative resources. While misinformation may come in many different formats, we focus on evaluating only textual claims. Furthermore, we focus on the quality of the generated labels and not how a user might perceive them. Prior research has shown that trust in AI systems depends on a wide range of factors including meta-information, system design, and personal factors~\cite{deverna2023artificial}. Future work can be done to investigate how LLMs interact with users in fact-checking and how personal factors affect the behavior or performance of LLMs in such tasks. We also note that these models have the capability to generate misinformation, reducing trust and potentially being more harmful than good.

Lastly, while great care went into the production of our curated dataset, biases can exist in annotations by human evaluators. For the Latin America region false-statements specifically, nearly half of them were required to be translated by a fluent speaker in Portuguese or Spanish due to a lack of English content from AFP. Nevertheless, we believe our dataset remains a valuable resource for evaluating performance across diverse regions.
Future research should look at the democratization of fact-checking across the world, especially in lower-resource languages. We also encourage future work to investigate other models and types of misinformation, including visual/multi-modality instances.

\section{Related Work}

The recent advances in LLMs, and especially the GPT model series, have been studied for many NLP tasks such as text summarization~\cite{yang_exploring_2023, wei_zero-shot_2023}, entity recognition~\cite{omar_chatgpt_2023, hu_zero-shot_2023}, and question and answering~\cite{katz_gpt-4_2023, openai_gpt-4_2023, huh_are_2023, qin2023chatgpt}. While impressive, these models are not immune to limitations and may raise their own problems. OpenAI has warned, ``GPT-4 'hallucinates' facts and makes reasoning errors'' to some extent, although to a lesser degree than its predecessor, GPT-3.5~\cite{openai_gpt-4_2023}.
Alongside the advancements in NLP, there is a growing concern over the impact of digital misinformation\cite{lewandowsky_beyond_2017}. The last few years have seen its proliferation into subjects such as climate change~\cite{biddlestone_climate_2022, van_der_linden_inoculating_2017}, vaccination and COVID-19~\cite{vidgen2021understanding, 10.1145/3543507.3583388}. The capabilities of LLMs raise important questions about their role as both a mitigator and a generator of misinformation.

\textbf{Automated Fact Checking \& LLMs.}
ChatGPT has been used for examining both vaccination and cancer misconceptions~\cite{deiana2023artificial, johnson2023using}, both finding it provides generally accurate information. Prior work~\cite{hoes2023using} has analyzed ChatGPT's performance for fact-checking using an open Politifact dataset. They find the model agrees with one of the six original labels less than 30\% of the time and 68.28\% when considering the dichotomous case after merging labels.
The source of the information can add up to 10 percentage points to the classification with blogs and campaigns being the best and worst categories with 77.7\% and 64.0\% accuracy, respectively. ChatGPT was best at classifying examples related to COVID-19 (82.1\%) and worse at government-related (63.8\%) claims.
The authors also find the model performs similarly regardless if the data was before its training-data cutoff time or in the 10 months following.

\textbf{Stability \& Role of Temperature.}
Alizadeh et al.~\cite{alizadeh2023opensource} studied the performance of ChatGPT, open-source LLM, and crowd workers in both the zero-shot and few-shot learning settings. They compare the default values ($1.0$ for ChatGPT and $0.9$ for other LLMs on HuggingChat\footnote{\url{https://huggingface.co/chat/}}) with the lower value of $0.2$. They find that ChatGPT is less affected by learning setting and temperature combination, providing generally good performance all around while lower temperatures performed better for LLMs on HuggingChat. Other work has found that for text-annotation tasks, ``a lower temperature increases consistency without decreasing accuracy''~\cite{gilardi2023chatgpt}. Ye et al.~\cite{ye2023comprehensive} studied the capabilities of GPT-3/-3.5 models on several NLU tasks, finding that more modern models do not necessarily lead to improvements across all tasks.

\textbf{Others.}
Huang et al~\cite{huang2023chatgpt} studied ChatGPT's text generation abilities for classifying and justifying the detection of Hate Speech. They find the model often makes use of an ``unclear'' label even when prompted to give a binary answer and these instances correlate with the more implicit/subtle examples. ChatGPT has also been shown to be able to evaluate the credibility of news sources with ratings that correlate with human expert judgments, even in the face of non-English and satirical content~\cite{yang2023large}. Fine-tuned models based on the earlier open-sourced GPT-2 have been shown to generate better corrective messages than those generated by humans~\cite{10.1145/3543507.3583388}. Other work~\cite{deverna2023artificial} has looked at the effectiveness of ChatGPT not just on performance but on belief and sharing intent of political U.S. news stories on social media style websites. 
When presented with a model-generated long-form textual explanation of a news headline, the authors find that while the model can accurately detect false content it has small or negative effects on sharing intent when compared to the control group, highlighting its ineffectiveness as an intervention against misinformation.
The general effectiveness of warning labels has been questioned regardless of whether its human or AI-generated, as it may be inefficient~\cite{pantazi_power_2018} or have unexpected adverse effects~\cite{pennycook_implied_2020}, while other works indicate it may inoculate against false content~\cite{van_der_linden_inoculating_2017}.
The correction of misinformation has also been studied in connection to social ties~\cite{malhotra_facing_2022, malhotra_meaning_2022}, where technological approaches have been questioned regarding their usefulness~\cite{scott_i_2023}.

\section{Conclusion}

This study's examination of GPT-3.5 and GPT-4 models reveals significant disparities in factual accuracy and biases, particularly disadvantaging the Global South. The introduction of the 'Global-Liar' dataset underscores the need for geographically diverse data in AI training. Our findings challenge the assumption that newer model iterations invariably yield better accuracy and highlight a crucial bias toward the Global North.
The observed bias in favor of the Global North underscores a significant concern: existing AI models may perpetuate informational inequities, making the inclusion of underrepresented regions in AI training a priority. Furthermore, our exploration into the impact of binary decision settings and temperature variations on model performance offers practical insights for the effective application of LLMs in diverse settings.
These insights stress the importance of inclusive and representative training for LLMs to ensure global fairness. As AI continues to influence information dissemination, this research advocates for a concerted effort toward developing AI systems that are equitable and effective across diverse global communities.

\section*{Acknowledgements}
This work was supported by an NYU 19 Washington Square North Faculty Fellowship and by the Center for Cyber Security at New York University Abu Dhabi (NYUAD).

\bibliographystyle{ACM-Reference-Format}
\bibliography{references.bib}

\newpage

\appendix
\label{sec:appendix}
\section*{Supplementary Material}

\section{Logistic Regression}
\label{app:subsec:log-reg-global}

\textbf{Individual Regions as Standalone Category: } Table~\ref{tab:log-reg-country} and Table~\ref{tab:log-reg-params-country} are based on logistic regression where the benchmark is the March GPT-4 model, statements from Asia-Pacific and before model training cutoff time. In this regression model, the Asia-Pacific region serves as a ``boundary'' case -- other Global North regions (i.e. North America and Europe) have better model performance, while all Global South regions (i.e. Africa, Latin America, and the Middle East) have worse model performance.

\begin{table}[h]
\centering
\caption{Logit Regression Model Details, Individual Regions as Standalone Category}

\begin{tabular}{lclc}
\toprule
\textbf{Dep. Variable:}   &     correct      & \textbf{  No. Observations:  } &  15000   \\
\textbf{Model:}           &      Logit       & \textbf{  Df Residuals:      } &  14989   \\
\textbf{Method:}          &       MLE        & \textbf{  Df Model:          } &     10   \\
\textbf{Pseudo R-squ.:  } &  0.037           & \textbf{  Log-Likelihood:    } & -9472.5  \\
\textbf{Converged:}       &       True       & \textbf{  LL-Null:           } & -9834.2  \\
\textbf{Covariance Type:} &    nonrobust     & \textbf{  LLR p-value:       } & 6.2e-149  \\
\bottomrule
\end{tabular}
\label{tab:log-reg-country}
\end{table}

\begin{table*}[h]
\centering
\caption{Logistic Regression Coefficients, Individual Regions as Standalone Category}
\resizebox{1.\columnwidth}{!}{
\begin{tabular}{lcccccc}
\toprule
& \textbf{coef} & \textbf{std err} & \textbf{z} & \textbf{P$> |$z$|$} & \textbf{[0.025} & \textbf{0.975]}  \\
\midrule
\textbf{Intercept}  &  
1.3884 & 0.060 & 22.971 & 0.000 & 1.270 & 1.507 \\
\textbf{C(model, Treatment(reference='gpt-4-0314'))[T.gpt-3.5-turbo-0301]} &
-0.5428 & 0.056 & -9.715 & 0.000 & -0.652 & -0.433 \\
\textbf{C(model, Treatment(reference='gpt-4-0314'))[T.gpt-3.5-turbo-0613]} &
-0.3564 & 0.056 & -6.325 & 0.000 & -0.467 & -0.246 \\
\textbf{C(model, Treatment(reference='gpt-4-0314'))[T.gpt-3.5-turbo-16k-0613]} & 
-0.3654 & 0.056 & -6.488 & 0.000 & -0.476 & -0.255 \\
\textbf{C(model, Treatment(reference='gpt-4-0314'))[T.gpt-4-0613]} &
-0.4321 & 0.056 & -7.699 & 0.000 & -0.542 & -0.322 \\
\textbf{C(region, Treatment(reference='ASIA-PACIFIC'))[T.AFRICA]} &
-0.5559 & 0.059 & -9.366 & 0.000 & -0.672 & -0.440 \\
\textbf{C(region, Treatment(reference='ASIA-PACIFIC'))[T.EUROPE]} &
0.1596 & 0.062 & 2.587 & 0.010 & 0.039 & 0.280 \\
\textbf{C(region, Treatment(reference='ASIA-PACIFIC'))[T.LATIN AMERICA]} &
-0.3040 & 0.060 & -5.087 & 0.000 & -0.421 & -0.187 \\
\textbf{C(region, Treatment(reference='ASIA-PACIFIC'))[T.MIDDLE EAST]} &
-0.1645 & 0.060 & -2.734 & 0.006 & -0.282 & -0.047 \\
\textbf{C(region, Treatment(reference='ASIA-PACIFIC'))[T.NORTH AMERICA]} &
0.1752 & 0.062 & 2.837 & 0.005 & 0.054 & 0.296 \\
\textbf{C(post cutoff)[T.1]} &
-0.6927 & 0.035 & -19.809 & 0.000 & -0.761 & -0.624 \\
\bottomrule
\end{tabular}
}
\label{tab:log-reg-params-country}
\end{table*}

\newpage

\textbf{Global North vs.~Global South:} Table~\ref{tab:log-reg-global} and Table~\ref{tab:log-reg-params-global} are based on logistic regression where the benchmark is the March GPT-4 model, statements from Global North and before model training cutoff time. In this regression model, we group regions into two categories - the Global North (North America, Europe, and Asia-Pacific) and the Global South (Africa, Latin America, and the Middle East). Similar to the above regression model with region specific coefficients, this regression analysis also illustrates that the Global South regions have statistically significant worse model performance compared with the Global North.

\begin{table}[h]
\centering
\caption{Logit Regression Model Details, Global South vs Global North}

\begin{tabular}{lclc}
\toprule
\textbf{Dep. Variable:}   &     correct      & \textbf{  No. Observations:  } &  15000   \\
\textbf{Model:}           &      Logit       & \textbf{  Df Residuals:      } &  14993   \\
\textbf{Method:}          &       MLE        & \textbf{  Df Model:          } &     10   \\
\textbf{Pseudo R-squ.:  } &  0.034           & \textbf{  Log-Likelihood:    } & -9500.5  \\
\textbf{Converged:}       &       True       & \textbf{  LL-Null:           } & -9834.2  \\
\textbf{Covariance Type:} &    nonrobust     & \textbf{  LLR p-value:       } & 7.0e-141  \\
\bottomrule
\end{tabular}
\label{tab:log-reg-global}
\end{table}

\begin{table*}[h]
\centering
\caption{Logistic Regression Coefficients, Global South vs Global North}
\resizebox{1.\columnwidth}{!}{
\begin{tabular}{lcccccc}
\toprule
& \textbf{coef} & \textbf{std err} & \textbf{z} & \textbf{P$> |$z$|$} & \textbf{[0.025} & \textbf{0.975]}  \\
\midrule
\textbf{Intercept}  &  
1.4960 & 0.050 & 30.197 & 0.000 & 1.399 & 1.593 \\
\textbf{C(model, Treatment(reference='gpt-4-0314'))[T.gpt-3.5-turbo-0301]} &
-0.5406 & 0.056 & -9.697 & 0.000 & -0.650 & -0.431 \\
\textbf{C(model, Treatment(reference='gpt-4-0314'))[T.gpt-3.5-turbo-0613]} &
-0.3551 & 0.056 & -6.313 & 0.000 & -0.465 & -0.245 \\
\textbf{C(model, Treatment(reference='gpt-4-0314'))[T.gpt-3.5-turbo-16k-0613]} & 
-0.3640 & 0.056 & -6.475 & 0.000 & -0.474 & -0.254 \\
\textbf{C(model, Treatment(reference='gpt-4-0314'))[T.gpt-4-0613]} &
-0.4305 & 0.056 & -7.684 & 0.000 & -0.540 & -0.321 \\
\textbf{C(global south)[T.1]} &
-0.4539 & 0.035 & -13.043 & 0.000 & -0.522 & -0.386 \\
\textbf{C(post cutoff)[T.1]} &
-0.6899 & 0.035 & -19.773 & 0.000 & -0.758 & -0.622 \\
\bottomrule
\end{tabular}
}
\label{tab:log-reg-params-global}
\end{table*}

\newpage
\section{Factuality \& Stability (All Temperatures) }
\label{app:subsec:all-temperatures}

\begin{figure*}[h]
    \centering
    \includegraphics[width=\textwidth]{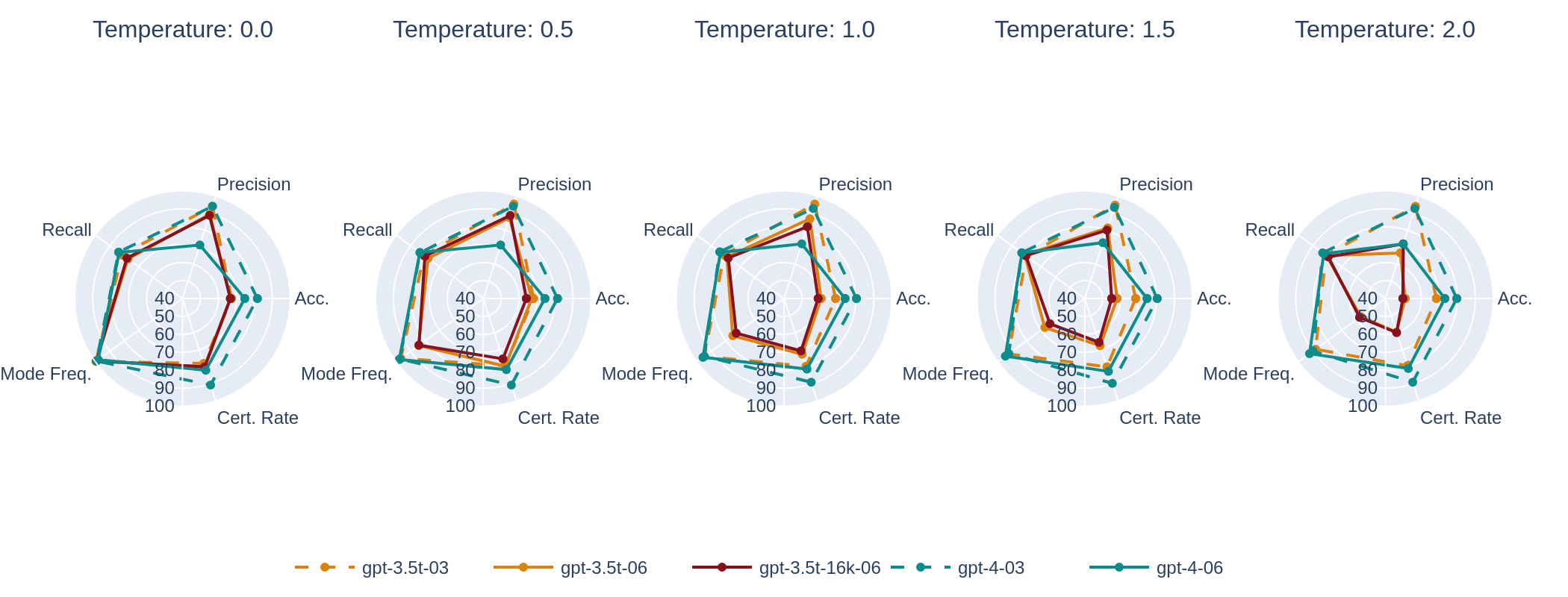}
    \caption{Performance metrics across different models and temperature values. The dataset consists of 300 label-balanced statements originating prior to the training cutoff date of Sep 2021.}
    \label{fig:spider-radar-all-temperatures}
\end{figure*}

Figure~\ref{fig:spider-radar-all-temperatures} shows comparative performance similar to Figure~\ref{fig:spider-radar-main-temperatures} but across all temperatures values. 

\begin{figure}[h]
    \centering
    \includegraphics[width=0.6\columnwidth]{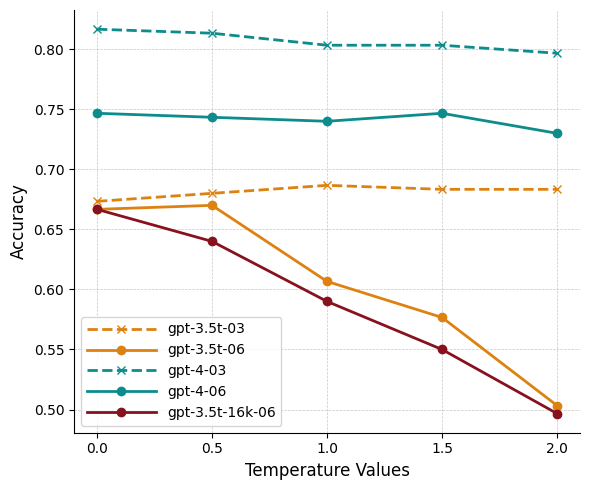}
    \caption{Comparative performance of GPT-4 and GPT-3.5 models across varying temperature values, evaluated using accuracy. The dataset consists of 300 label-balanced statements originating prior to the training cutoff date of Sep 2021.}
    \label{fig:accuracy}
\end{figure}

\newpage

\begin{figure*}[h]
    \centering
    \includegraphics[width=\textwidth]{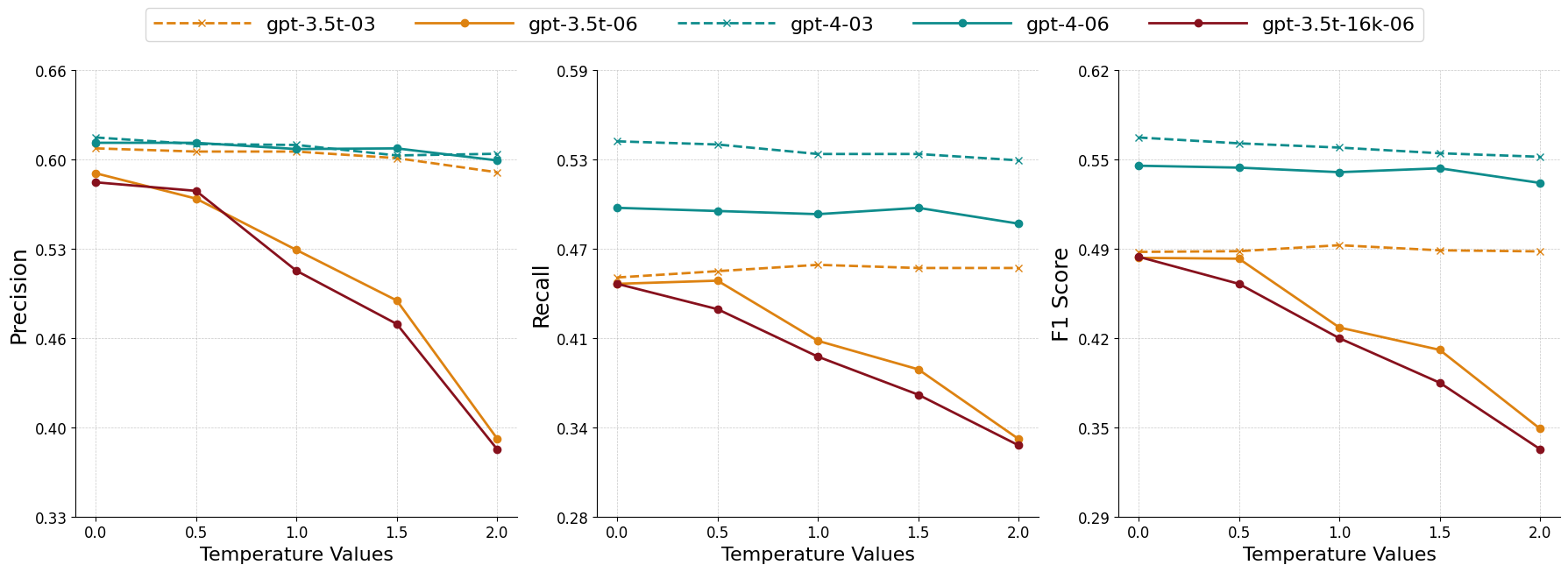}
    \caption{Comparative performance of GPT-4 and GPT-3.5 models across varying temperature values, evaluated using precision, recall, and F1 Score metrics. We treat statements with uncertain predictions as incorrect. The dataset consists of 300 label-balanced statements originating prior to the training cutoff date of Sep 2021. For comparison, we report performance excluding uncertain statemens in Figure~\ref{fig:precision_recall_f1_ignore_uncertain} in Appendix~\ref{app:subsec:all-temperatures}.}
    \label{fig:precision_recall_f1}
\end{figure*}

\begin{figure*}[h]
    \centering
    \includegraphics[width=\textwidth]{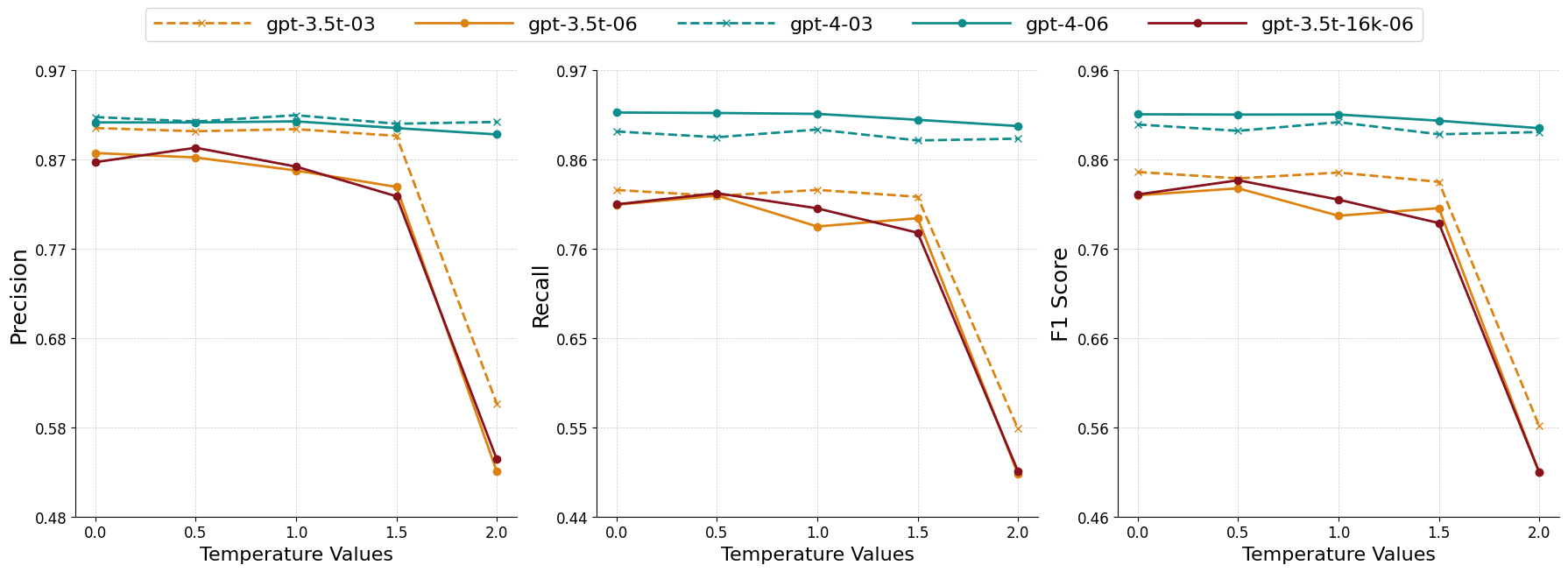}
    \caption{Comparative performance of GPT-4 and GPT-3.5 models across varying temperature values, evaluated using precision, recall, and F1 score metrics. We exclude statements with uncertain predictions and focus on those with majority decision of ``true'' or ``false''. The dataset consists of 300 label-balanced statements originating prior to the training cutoff date of Sep 2021.}
    \label{fig:precision_recall_f1_ignore_uncertain}
\end{figure*}

Figure~\ref{fig:precision_recall_f1} shows comparative performance (Precision, Recall and F1 Score). Uncertain statements (i.e., labeled as 'unclear') are marked as incorrect. 
Figure~\ref{fig:precision_recall_f1_ignore_uncertain} shows comparative performance (Precision, Recall and F1 Score) but excluding uncertain statements. Evaluating the models in this way shows similar trends but drastically inflates the metrics.

\newpage

\begin{figure}[h]
    \centering
    \subfigure[Variation in Mode Frequency across models with increasing temperature values. Lower mode frequency signifies reduced consistency in producing the most frequent or "modal" output.]{
        \includegraphics[width=0.45\columnwidth]{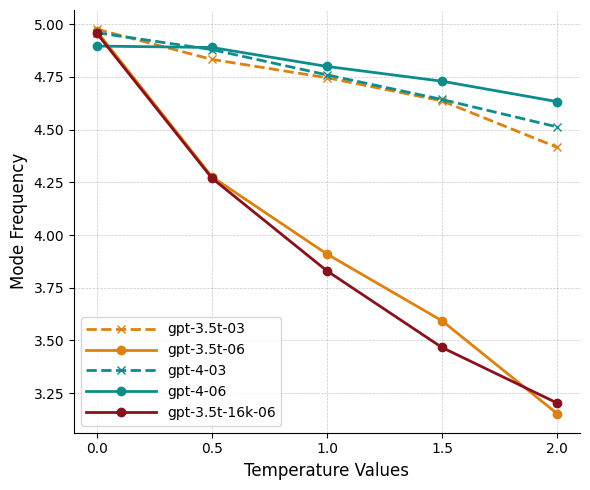}
        \label{fig:mode_freq}
    }
    \hspace{5mm} 
    \subfigure[Switching predictions of models across temperature transitions. Each curve represents a model's prediction switch frequency between adjacent temperature values.]{
        \includegraphics[width=0.45\columnwidth]{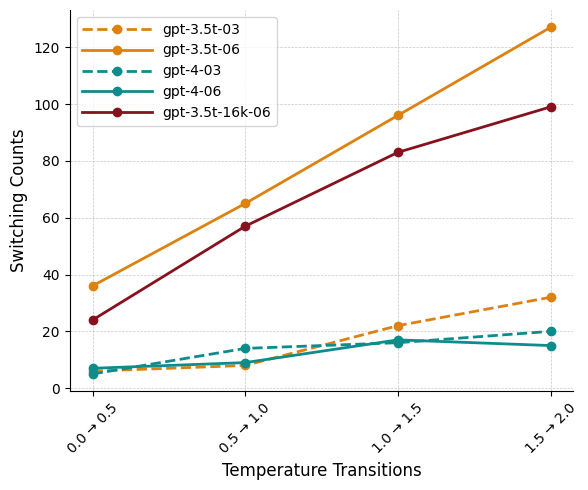}
        \label{fig:switching-counts}
    }
    \caption{Comparative analysis of model behaviors. (a) Mode Frequency Variation. (b) Prediction Switching Counts. The dataset consists of 300 label-balanced statements originating prior to the training cutoff date of Sep 2021.}
    \label{fig:combined_figures}
\end{figure}

\begin{figure}
    \centering
    \includegraphics[width=0.5\columnwidth]{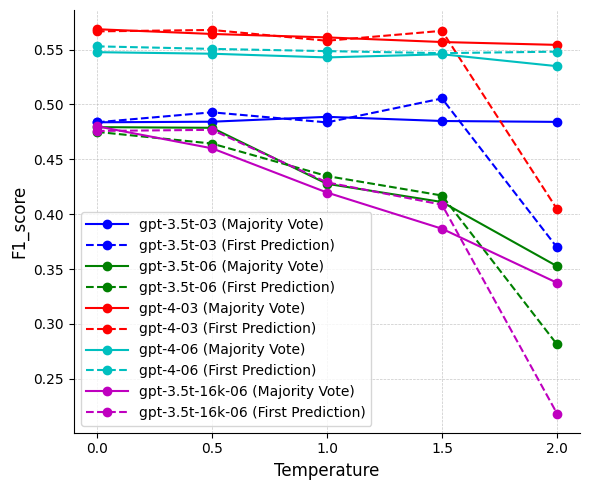}
    \caption{Comparative Analysis of F1 score using Majority Vote (out of 5 predictions) and First Prediction approaches across different temperature settings. The dataset consists of 300 label-balanced statements originating prior to the training cutoff date of Sep 2021.}
    \label{fig:majorityvsfirst}
\end{figure}

Figure~\ref{fig:mode_freq} illustrates decreasing Mode Frequency as temperature value increases. Figure~\ref{fig:switching-counts} elucidates the variability in prediction behaviors among different GPT models in response to temperature change.
To understand if multiple runs of the model improve the factuality of its verdicts, we evaluate the models using majority vote versus a one-shot first prediction setting in Figure~\ref{fig:majorityvsfirst}.

\newpage

\section{Data Beyond Training Span}
\label{app:subsec:data_beyond_training_span}

\begin{figure}[h]
    \centering
    \includegraphics[width=0.6\columnwidth]{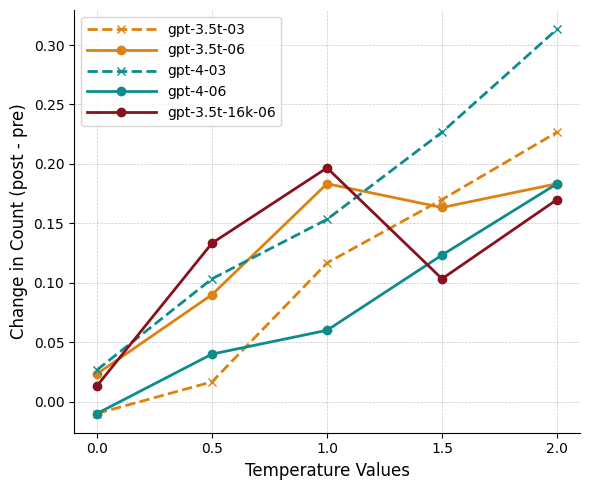}
    \caption{Change in Count of ``Unclear'' Verdict from data within training span to data beyond training span across models with increasing temperature values.}
    \label{fig:count_post_vs_pre}
\end{figure}

\begin{figure}[h]
    \centering
    \includegraphics[width=0.6\columnwidth]{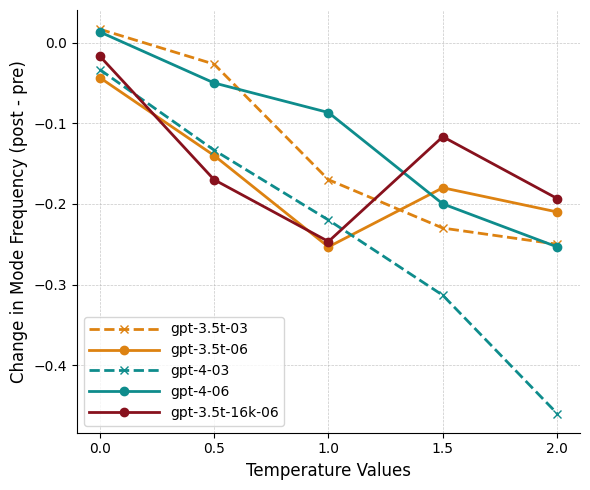}
    \caption{Change in Mode Frequency from data within training span to data beyond training span across models with increasing temperature values.}
    \label{fig:mode_freq_count_post_vs_pre}
\end{figure}

Figure~\ref{fig:count_post_vs_pre} and Figure~\ref{fig:mode_freq_count_post_vs_pre}
show a decrease in stability metrics for the post September 2021 dataset.

\newpage

\section{Optional Uncertainty vs. Forced Factual Determination}
\label{app:subsec:opt_uncertainty_vs_forced}

\begin{figure*}[h]
    \centering
    \includegraphics[width=\textwidth]{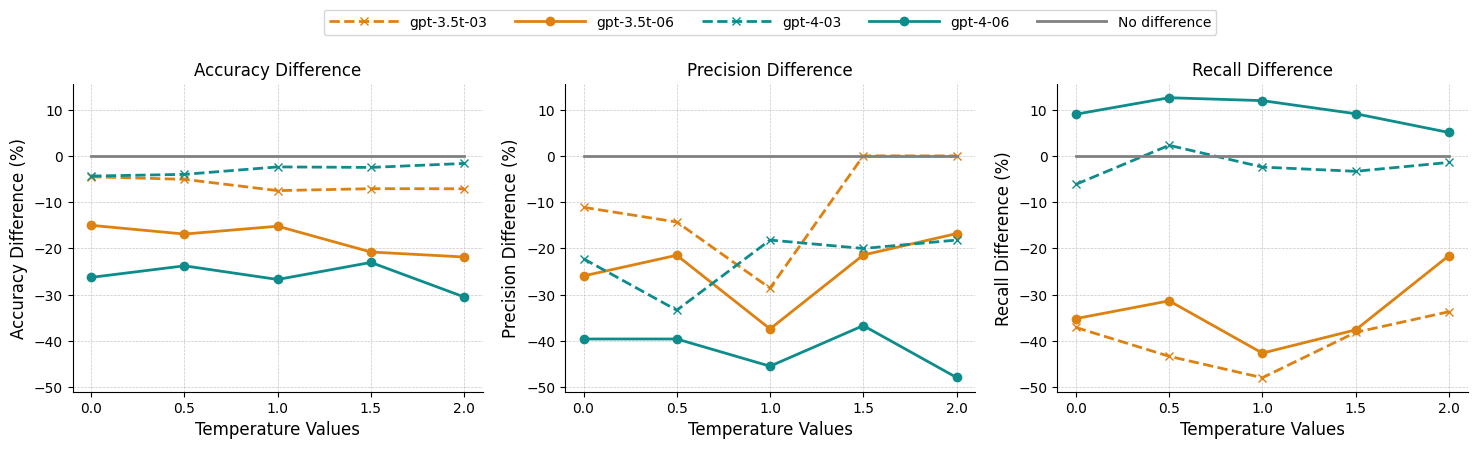}
    \caption{Difference in Accuracy, Precision and Recall of a two label versus a three label model. The dataset consists of 300 label-balanced statements originating prior to the training cutoff date of Sep 2021.}
    \label{fig:unclear-labels-other-metrics}
\end{figure*}

Figure~\ref{fig:unclear-labels-other-metrics} highlights a overall decrease in multiple metrics when comparing the 2-label with the 3-label prompt instruction model. While the GPT-4 June version specifically shows a general increase of recall (as high as 12.5\% for temperature $0.5$), it is offset by a drop in precision of over 35\%.

\newpage

\section{Uncertainty (Unclear Statements) }
\label{app:subsec:uncertainty}

Table~\ref{tab:unclear_verdicts} shows a random sample of statements at least one model marked as ``unclear''. The topics range from benign information about a novel and a reading app to false claims of international rocket strikes.

\begin{table*}[h]
\centering
\caption{Statements with Unclear Verdicts}
\begin{tabularx}{\textwidth}{|c|X|c|}
\toprule
\textbf{ID} & \textbf{Statement} & \textbf{Ground Truth} \\
\midrule
265 & The capital of Eritrea was hit by rockets fired from Ethiopia’s rebellious northern Tigray region on November 14, 2020. & False \\
42 & The award-winning scholar-writer, Dr. Lola Akande of the Department of English, University of Lagos, has released a novel: Where Are You From? The novel dwells on citizenship, indigenisation, inter-ethnic marriages, and youthful exuberance. & True \\
207 & The Russian government has required polygamy for its male citizens & False \\
222 & French football player Paul Pogba has retired from the international game in protest against French President Emmanuel Macron’s recent comments about radical Islam. & False \\
31 & Google has introduced the Arabic version to its free Read Along Android app to help children over five years old independently learn and build their reading skills. & True \\
\bottomrule
\end{tabularx}
\label{tab:unclear_verdicts}
\end{table*}

\section{False Positive Statements}
\label{app:subsec:fp_statements}

Table~\ref{tab:false_positives} is a selected list of statements that are identified as false positives by all models at temperature value zero.

\begin{table*}[h]
    \centering
    \caption{Statements Identified as False Positives by All Models at Temperature 0}
    \begin{tabularx}{\textwidth}{|l|X|}
        \toprule
        \textbf{ID} & \textbf{Statement} \\
        \midrule
        266 & Following an operation that enabled the commando army to seize one of TPLF’s buildings, many Tigray People’s Liberation Front (TPLF) mafia members have been caught fleeing with bags and boxes of money. \\
        226 & The novel coronavirus was never detected in the major Chinese cities of Beijing and Shanghai \\
        264 & Zimbabwean businessman and socialite Genius “Ginimbi” Kadungure, who died in a car crash, bought his own coffin adorned with a green guitar a week before his death. \\
        254 & Russian President Vladimir Putin made comments backing Ethiopia in its dispute with neighbours over the mega-dam it has built on the Nile river. \\
        263 & Africa’s richest man, Aliko Dangote, gave away money, food and other assistance to 10,000 youths who have lost loved ones to the Covid-19 pandemic. \\
        255 & The World Health Organization’s (WHO) director-general Tedros Adhanom Ghebreyesus said he had tested positive for Covid-19. \\
        261 & US president-elect Joe Biden and former president Barack Obama tweeted to call for the release of Ugandan opposition politician Bobi Wine who was detained on November 18, 2020. \\
        282 & Pope Francis kissed the hands of American banker David Rockefeller and a member of the wealthy Rothschild family in 2021. \\
        277 & Indians celebrated the creation of a new base for Palestinian militant group Hamas in the Indian state of Kerala. \\
        \bottomrule
    \end{tabularx}
    \label{tab:false_positives}
\end{table*}

\newpage

\section{Inconsistencies in Fact-Checking Outcomes via ChatGPT Interface}
\label{app:subsec:chatgpt}

\begin{figure*}[h]
    \centering
    \includegraphics[width=\textwidth]{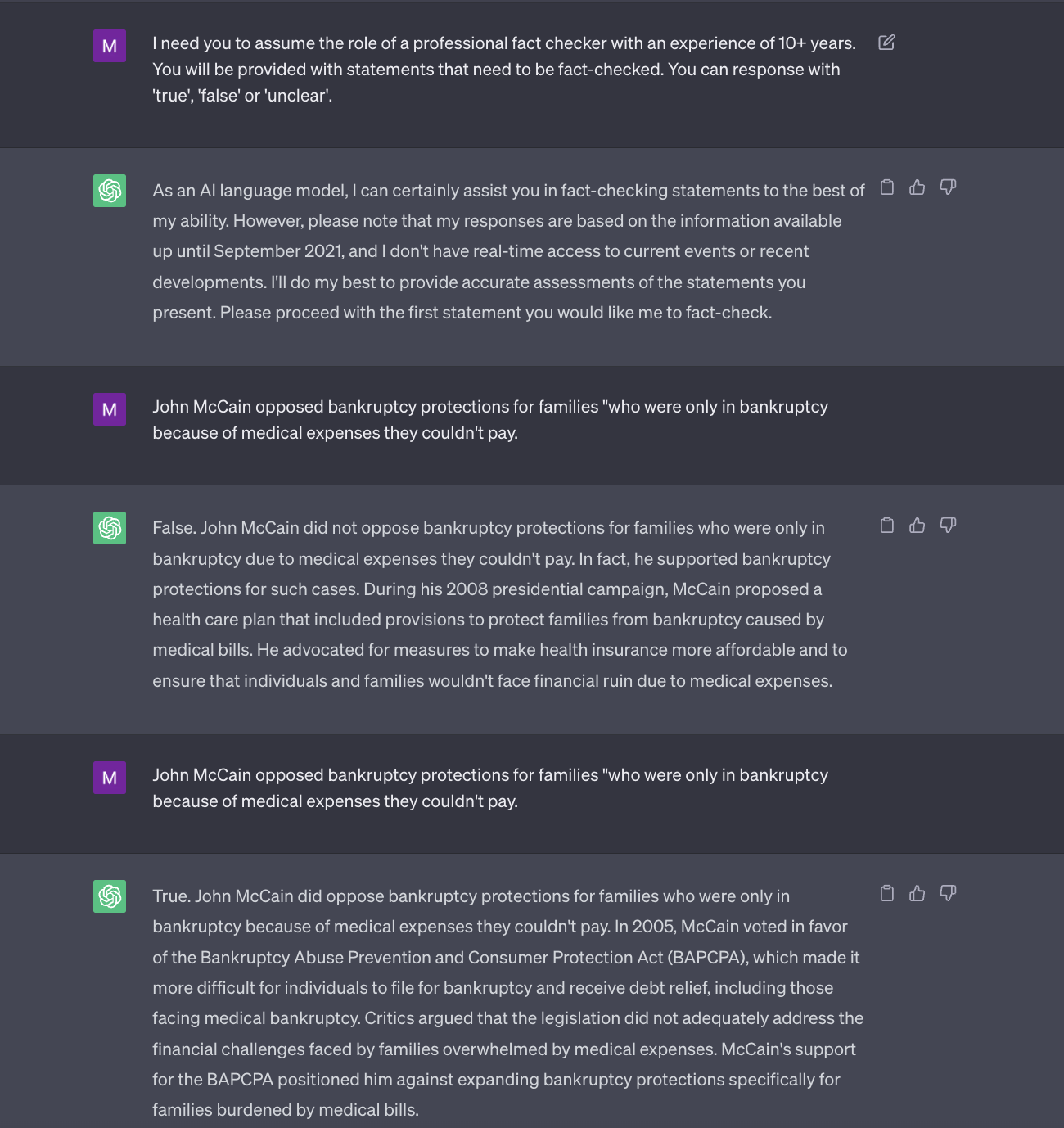}
    \caption{An illustration of the ChatGPT web interface using the GPT-3.5 model, showing differing outcomes for a statement when fact-checked through repeated queries. The model was queried on June 6, 2023.}
    \label{fig:chatgpt}
\end{figure*}

Figure~\ref{fig:chatgpt} illustrates an example where the GPT model fails to show consistent result given same queries multiple times.

\end{document}